\documentclass[journal]{IEEEtran}

\usepackage{amsmath}
\usepackage{amssymb}
\usepackage{booktabs}
\usepackage{multirow}
\usepackage{enumitem}

\usepackage{amsthm,amsmath,amssymb}
\usepackage{mathrsfs}
\usepackage{stfloats}
\usepackage[pagebackref,breaklinks,colorlinks]{hyperref}

\usepackage[capitalize]{cleveref}
\ifCLASSINFOpdf
  \usepackage[pdftex]{graphicx}
  \DeclareGraphicsExtensions{.pdf,.jpeg,.png}
\else
  \usepackage[dvips]{graphicx}
  \DeclareGraphicsExtensions{.eps}
\fi
\begin{document}
%
\title{U-shape Transformer\\ for Underwater Image Enhancement}
%
%
%

\author{Lintao~Peng,
        Chunli~Zhu,
        and~Liheng~Bian*
\thanks{L. Peng, C. Zhu and L. Bian are with the Advanced Research Institute of Multidisciplinary Science \& School of Information and Electronics, Beijing Institute of Technology, Beijing, China. Correspondence to L. Bian: bian@bit.edu.cn.}}

%
%

\markboth{Journal of \LaTeX\ Class Files,~Vol.~14, No.~8, August~2015}%
{Shell \MakeLowercase{\textit{et al.}}: Bare Demo of IEEEtran.cls for IEEE Journals}
%



\maketitle

\begin{abstract}
The light absorption and scattering of underwater impurities lead to poor underwater imaging quality. The existing data-driven based underwater image enhancement (UIE) techniques suffer from the lack of a large-scale dataset containing various underwater scenes and high-fidelity reference images. Besides, the inconsistent attenuation in different color channels and space areas is not fully considered for boosted enhancement. In this work, we built a large scale underwater image (LSUI) dataset, which covers more abundant underwater scenes and better visual quality reference images than existing underwater datasets. The dataset contains 4279 real-world underwater image groups, in which each raw image's clear reference images, semantic segmentation map and medium transmission map are paired correspondingly. We also reported an U-shape Transformer network where the transformer model is for the first time introduced to the UIE task. The U-shape Transformer is integrated with a channel-wise multi-scale feature fusion transformer (CMSFFT) module and a spatial-wise global feature modeling transformer (SGFMT) module specially designed for UIE task, which reinforce the network's attention to the color channels and space areas with more serious attenuation. Meanwhile, in order to further improve the contrast and saturation, a novel loss function combining RGB, LAB and LCH color spaces is designed following the human vision principle. The extensive experiments on available datasets validate the state-of-the-art performance of the reported technique with more than 2dB superiority. The dataset and demo code are available on \url{https://lintaopeng.github.io/_pages/UIE%20Project%20Page.html}.
\end{abstract}


%
\IEEEpeerreviewmaketitle

\section{Introduction}
%
%
%
%
\IEEEPARstart{U}{nderwater} Image Enhancement (UIE) technology \cite{yang2019depth,sahu2014survey} is essential for obtaining underwater images and investigating the underwater environment, which has wide applications in ocean exploration, biology, archaeology, underwater robots \cite{Islam2020FUnIE} and among other fields. However, underwater images frequently have problematic issues, such as color casts, color artifacts and blurred details \cite{schettini2010underwater}. Those issues could be explained by the strong absorption and scattering effects on light, which are caused by dissolved impurities and suspended matter in the medium (water). Therefore, UIE-related innovations are of great significance for improving the visual quality and merit of images in accurately understanding the underwater world.
\begin{figure}[t] 
	\centering
	\includegraphics[width=1\linewidth]{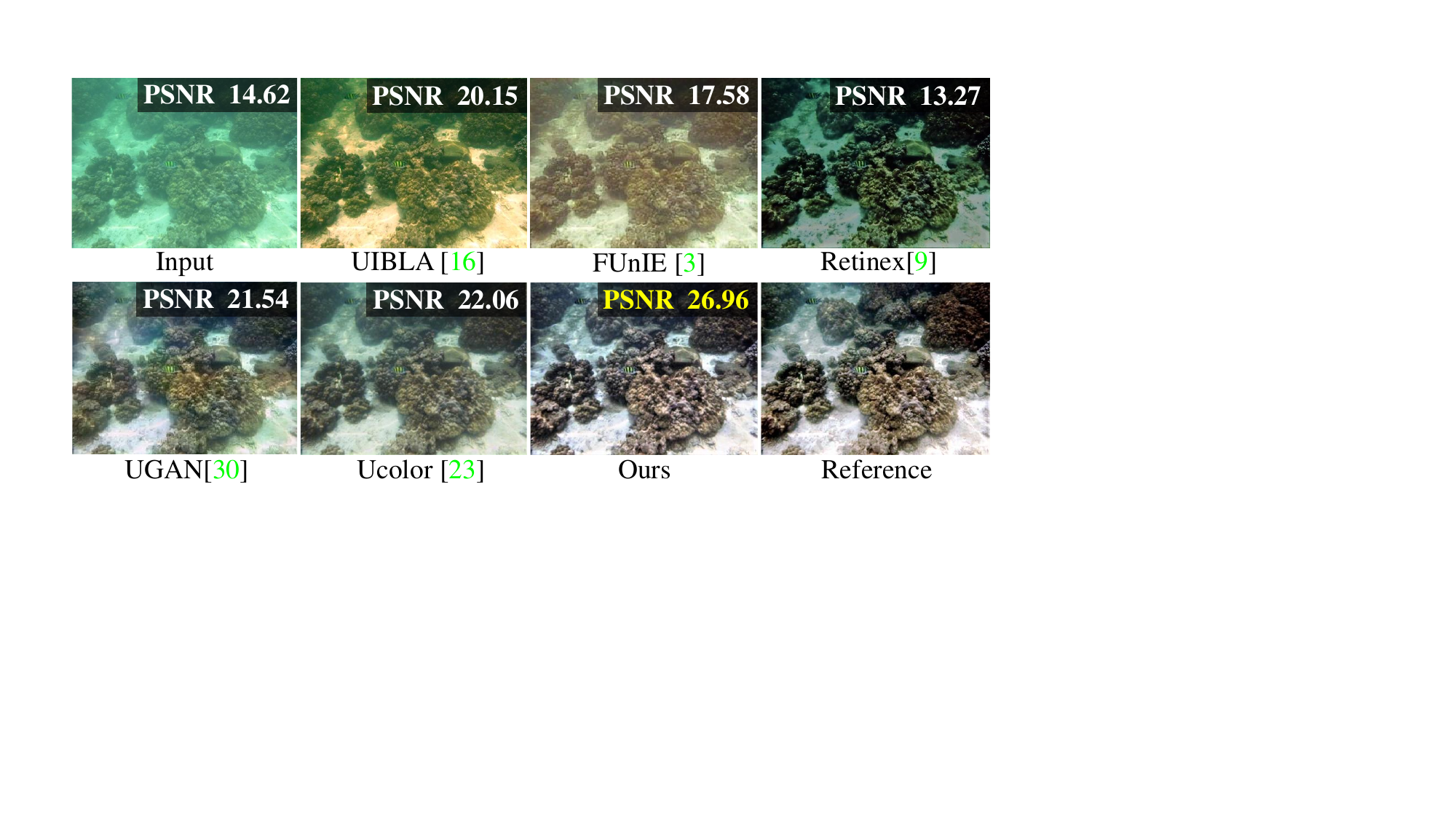}
	\caption{Compared with the existing UIE methods, the image produced by our U-shape Transformer has the highest PSNR\cite{korhonen2012peak} score and best visual quality.}
	\label{figure:1}
\end{figure}

In general, the existing UIE methods could be categorized into three types, which are physical model-based, visual prior-based and data-driven methods, respectively. Among them, visual prior-based UIE methods \cite{Ancuti2012fusion,Li2016Histogram,Ghani2014RayleighDistribution,Fu2014Retinexbased,huang2018RGHS,Iqbal2010UIM} mainly concentrated on improving the visual quality of underwater images by modifying pixel values from the perspectives of contrast, brightness and saturation. Nevertheless, the ignorance of the physical degradation process limits the improvement of enhancement quality. In addition, physical-model based UIE methods \cite{He2009DCP,Drews2013UDCP,Drews2016RBSI,Wang2018aacp,peng2017underwater,Li2016miniinfor,Chiang2011WCD,galdran2015automatic,Li2016miniop} mainly focus on the accurate estimation of medium transmission. With the estimated medium transmission and other key underwater imaging parameters such as the homogeneous background light, a clean image can be obtained by reversing a physical underwater imaging model. However, the performance of physical model-based UIE is restricted to complicated and diverse real-world underwater scenes. That is because, (1)\emph{ model hypothesis is not always plausible with complicated and dynamic underwater environment}; (2)\emph{ evaluating multiple parameters simultaneously is challenging.} More recently, as to the data-driven methods \cite{guo2019mutiscaleGAN,Islam2020FUnIE,li2017watergan,Li2021UnderwaterIE,li2018cyclegan,Yang2011LDCP,Akkaynak2019seathru,Fu2017twostep,Li2019UIEB,song2018rapid,Fabbri2018UGAN,uplavikar2019UIEDAL,schettini2010underwater}, which could be regarded as deep learning technologies in UIE domain, exhibit impressive performance on UIE task. However, the existing underwater datasets more-or-less have the disadvantages, such as a small number of images, few underwater scenes, or even not real-world scenarios, which limits the performance of the data-driven UIE method. Besides, the inconsistent attenuation of the underwater images in different color channels and space areas have not been unified in one framework.

In this work, we first built a large scale underwater image (LSUI) dataset, which covers more abundant underwater scenes (water types, lighting conditions and target categories) and better visual quality reference images than existing underwater datasets \cite{Liu2020RealWorldUE,Li2019UIEB,Akkaynak2019seathru,li2017watergan}. The dataset contains 4279 real-world underwater images, and the corresponding clear images are generated as comparison references. We also provide the semantic  segmentation  map and medium  transmission  map for each image.  Furthermore, with the prior knowledge that the attenuation of different color channels and space areas in underwater images is inconsistent, we designed a channel-wise multi-scale feature fusion transformer (CMSFFT) and a spatial-wise global feature modeling transformer (SGFMT) based on the attention mechanism, and embedded them in our U-shape Transformer which is designed based on \cite{isola2017image}.
Moreover, according to the color space selection experiment and \cite{Xiaocong2019,Li2021UnderwaterIE}, we designed a multi-color space loss function including RGB, LAB and LCH color space. Fig. \ref{figure:1} shows the result of our UIE method and some comparison UIE methods, and the main contributions of this paper can be summarized as follows:
\begin{itemize}
	\item We reported a novel U-shape Transformer dealing with the UIE task, in which the designed channel-wise and spatial-wise attention mechanism based on transformer enables to effectively remove color artifacts and casts.
	\item We designed a novel multi-color space loss function combing the RGB, LCH and LAB color-space features, which further improves the contrast and saturation of output images.
	\item We released a large-scale dataset containing 4279 real underwater images and the corresponding high-quality reference images, semantic segmentation maps, and medium transmission maps, which facilitates further development of UIE techniques.
\end{itemize}

\section{Related work}
\subsection{UIE Methods}
UIE is an indispensable step to improve the visual quality of recorded underwater images. 
A variety of methods have been proposed and can be categorized as visual prior, physical models, and data-driven methods.

\textbf{UIE methods based on visual prior.} This approach aims to restore a clear underwater image by modifying its pixel value. Typical methods involve:
1) \textit{Modify the pixel value with single metric.} Such as contrast adjustment \cite{hitam2013contrast}, histogram equalization \cite{hitam2013contrast}, and white balance \cite{Iqbal2010UIM}. For instance, Hitam et al. \cite{hitam2013contrast} used contrast adjustment and adaptive histogram equalization methods in RGB color space and HSV color space to enhance the contrast of underwater images and reduce noise. 
2) \textit{Modify the pixel value with multiple metrics.} For example, fusion-based methods, which exhibit the final enhancement image via the weighted fusion of multiple traditional UIE methods. For example, Fang et al. \cite{Ancuti2012fusion} first applied white balance and global contrast adjustments to enhance underwater images, and then the two enhancement results are combined into one image by weighted addition to obtain the final enhanced underwater image. 
3) \textit{Retinex based UIE methods}.
Fu et al. \cite{Fu2014Retinexbased} proposed a retinex model-based UIE method including color correction, layer decomposition and enhancement. Furthermore, Zhang et al. \cite{ZHANG2017Retinex} proposed a multi-scale UIE method based on the retinex model.

The way of modifying pixel value has the inherent advantage of improving the contrast and saturation of the raw underwater image. However, as visual prior neglected the inconsistent attenuation degree of underwater images in varied color channels and space areas, it performs not well on real underwater images with complex underwater environments.

\textbf{UIE methods based on physical models.} 
This approach regard UIE as a problem of inversion, and researchers usually enhance underwater images based on the following three steps, 1) establishing the prior conditions of the hypothetical physical imaging model; 2) estimating the key parameters; 3) reversing the degradation process of the underwater imaging process to obtain a clear image.


Prior is the basis of the physical model based UIE, in which existing work includes underwater dark channel priors \cite{Drews2013UDCP}, attenuation curve priors \cite{Wang2018aacp}, fuzzy priors \cite{Chiang2011WCD} and minimum information priors \cite{Li2016miniop}, etc. Early-stage
research enhanced the underwater image by modifying the dark channel prior (DCP) \cite{Drews2013UDCP} algorithm. Chiang et al.\cite{Chiang2011WCD} restored the underwater image by combining the DCP with the wavelength compensation algorithm. Drews Jr et al. \cite{Drews2013UDCP} proposed an underwater dark channel prior algorithm (UDCP) based on the priori that the red channel in the underwater image is more attenuated. 
Carlevaris Bianca et al. \cite{Carlevaris2010initial} used the attenuation difference prior between the three color channels in RGB color space to predict the transmission characteristics of the underwater scene, which feasibility is basically due to red light generally decays faster than green and blue light.
In addition,  Peng et al. \cite{peng2017underwater} proposed a depth map estimate method for underwater scenes based on the intrinsic characteristics of underwater image blurriness and light absorption to effectively recover underwater images. Li et al. \cite{Li2016Histogram} integrate the minimum information loss and histogram distribution prior
for depth estimation to recover underwater images. 

This branch of UIE methods could achieve satisfactory results only when underwater scenes are in accordance with the selected physical imaging model. Therefore, the manually established priors restrain the model's robustness and scalability under the complicated and varied circumstances. Moreover, as the underwater physical imaging model does not taken human eye's perception characteristics into account, the visual quality of the restored images are of poor presentation effect. In recent years, underwater physical imaging models are gradually utilized in combination with data-driven methods\cite{Li2021UnderwaterIE}.





\textbf{Data-driven UIE methods.} 
Current data-driven UIE methods can be divided into two main technical routes,  (1) \emph{designing an end-to-end module;} (2) \emph{utilizing deep models directly to estimate physical parameters, and then restore the clean image based on the degradation model.}
To alleviate the need for real-world underwater paired training data, Li et al. \cite{li2017watergan} proposed a WaterGAN to generate underwater-like images from in-air images and depth maps in an unsupervised manner, in which the generated dataset is further used to train the WaterGAN. Moreover,  \cite{li2018cyclegan} exhibited a weakly supervised underwater color transmission model based on CycleGAN \cite{Zhu2017CycleGAN}. Benefiting from the adversarial network architecture and multiple loss functions, that network can be trained using unpaired underwater images, which refines the adaptability of the network model to underwater scenes. However, images in the training dataset used by the above methods are not matched real underwater images, which leads to limited enhancement effects of the above methods in diverse real-world underwater scenes. 
Recently, Li et al. \cite{Li2019UIEB}  proposed a gated fusion network named WaterNet, which uses gamma-corrected images, contrast-improved images, and white-balanced images as the inputs to enhance underwater images. Yang et al. \cite{yang2020underwater} proposed a conditional generative adversarial
network (cGAN) to improve the perceptual quality of underwater images.

The methods mentioned above usually use existing deep neural networks for general purposes directly on UIE tasks and neglect the unique characteristics of underwater imaging. For example, \cite{li2018cyclegan} directly used the CycleGAN \cite{Zhu2017CycleGAN} network structure, and \cite{Li2019UIEB} adopted a simple multi-scale convolutional network. Other models such as UGAN \cite{Fabbri2018UGAN},WaterGAN \cite{li2017watergan} and cGAN \cite{yang2020underwater}, still inherited the disadvantage of GAN-based models, which produces unstable enhancement results. In addition, Ucolor \cite{Li2021UnderwaterIE} combined the underwater physical imaging model and designed a medium transmission guided model to reinforce the network's response to areas with more severe quality degradation, which could improve the visual quality of the network output to a certain extent. However, physical models sometimes failed with varied underwater environments.

From above, our proposed network aims at generating high visual quality underwater images by properly accounting the inconsistent attenuation characteristics of underwater images in different color channels and space areas.

\subsection{Underwater Image Datasets}
The sophisticated and dynamic underwater environment results in extreme difficulties in the collection of matched underwater image training data in real-world underwater scenes. 
Present datasets can be classified into two types, they are,
(1) Non-reference datasets. Liu et al. \cite{Liu2020RealWorldUE} proposed the RUIE dataset, which encompasses varied underwater lighting, depth of field, blurriness and color cast scenes. Akkaynak et al. \cite{Akkaynak2019seathru} published a non-reference underwater dataset with a standard color comparison chart. Those datasets, however, cannot be used for end-to-end training for lacking matched clear reference underwater images. 
(2) Full-reference datasets. Li et al. \cite{li2017watergan} presented an unsupervised network dubbed WaterGAN to produce underwater-like images using in-air images and depth maps. Similarly, Fabbri et al. \cite{Fabbri2018UGAN} used CycleGAN to generate distorted images from clean underwater images based on weakly-supervised distribution transfer. However, these methods rely heavily on training samples, which is easy to produce artifacts that are out of reality and unnatural. 
Li et al. \cite{Li2019UIEB} constructed a real UIE benchmark UIEB, including 890 images pairs, in which reference images were hand-crafted using the existing optimal UIE methods. Although those images are authentic and reliable, the number, content and coverage of underwater scenes are limited. In contrast, our LSUI dataset contains 4279 real-world underwater images with more abundant underwater scenes (water types, lighting conditions and target categories) than existing underwater datasets \cite{Liu2020RealWorldUE,Li2019UIEB,Akkaynak2019seathru,li2017watergan}, and the corresponding clear images are generated as comparison references. We also provide the semantic  segmentation  map and medium  transmission  map for each raw underwater image.

\subsection{Transformers}
Although CNN-based UIE methods \cite{Li2019UIEB,Islam2020FUnIE,Fabbri2018UGAN,uplavikar2019UIEDAL,Li2021UnderwaterIE} achieved significant improvement compared with traditional UIE methods. There are still two aspects that limit its further promotion, (1)\emph{ uniform convolution kernel is not able to characterize the inconsistent attenuation of underwater images in different color channels and spatial regions;} (2) \emph{the CNN architecture concerns more on local features, while ineffective for long-dependent and global feature modeling.}

Recently, transformer \cite{vaswani2017attention} has gained more and more attention, its content-based interactions between image content and attention weights can be interpreted as spatially varying convolution, and the self-attention mechanism is efficient at modeling long-distance dependencies and global features. Benefiting from these advantages, transformers have shown outstanding performance in several vision tasks \cite{dosovitskiy2020image,liu2021swin,zamir2021restormer,Zhao_2021_ICCV}. Compared with previous CNN-based UIE networks, our CMSFFT and SGFMT modules designed based on the transformer can guide the network to pay more attention to the more serious attenuated color channels and spatial areas. Moreover, by combining CNN with transformer, we achieve better performance with a relatively small amount of parameters.

\section{Proposed dataset and method}

\subsection{LSUI Dataset}
\begin{figure}[h]
	\centering
	\includegraphics[width=1\linewidth]{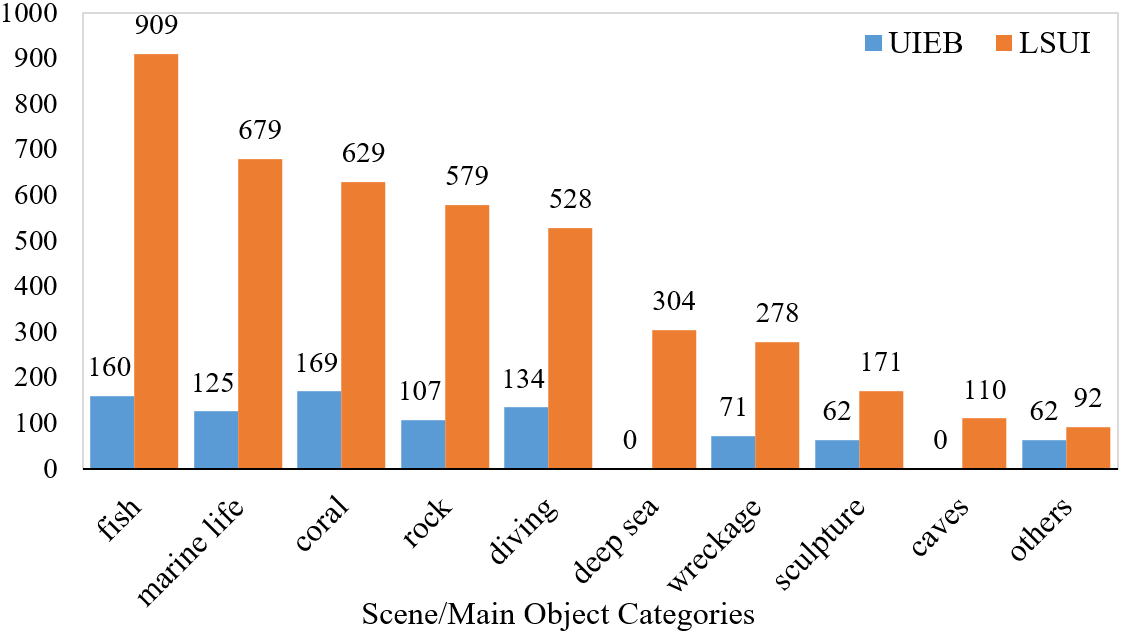}
	\caption{Statistics of our LSUI dataset and the existing underwater dataset UIEB \cite{Li2019UIEB}.}
	\label{figure:Statistics}
\end{figure}

\begin{figure*}[h]
	\centering
	\includegraphics[width=1\linewidth]{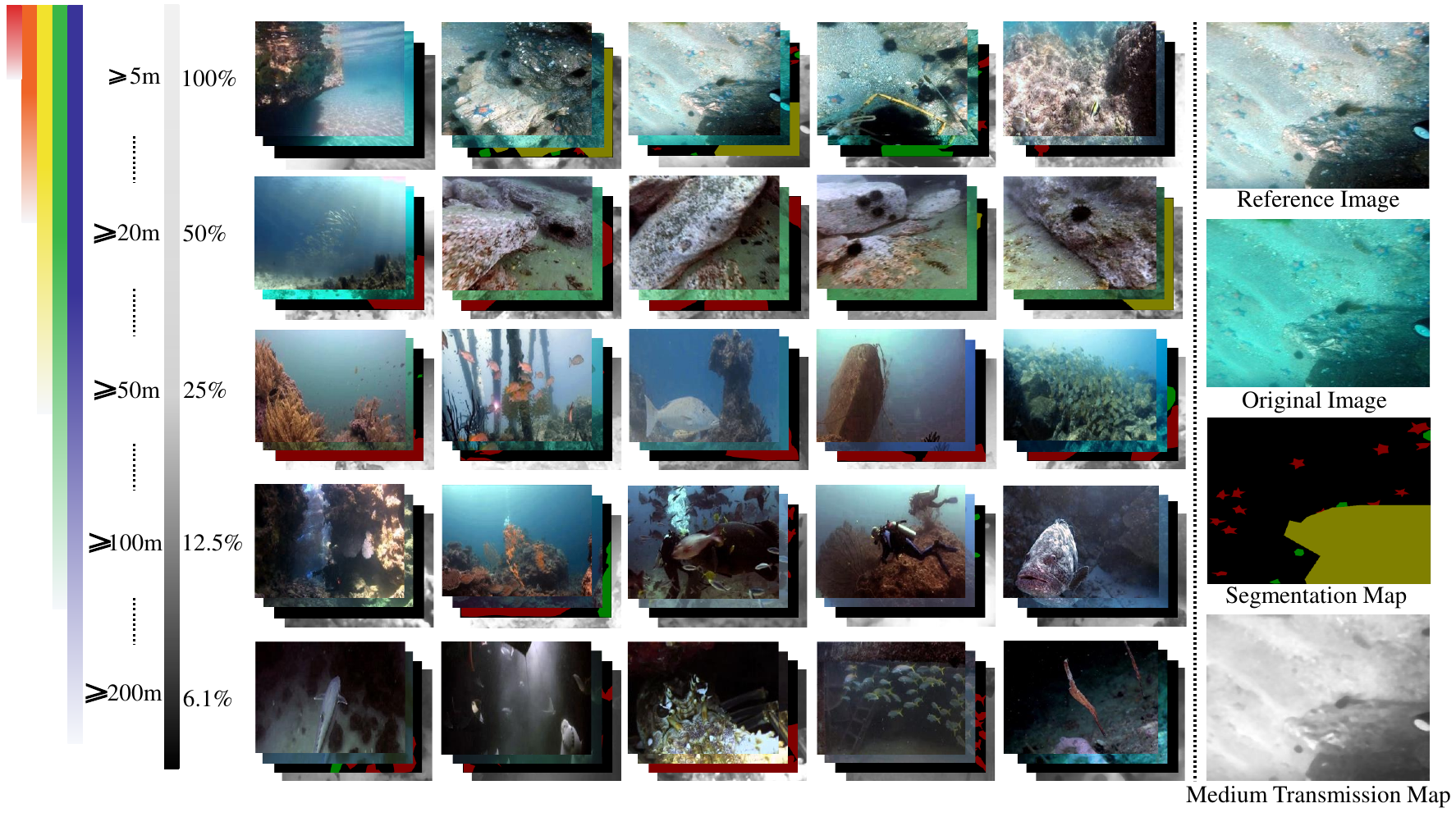}
	\caption{Example images in the LSUI dataset. Our LSUI dataset contains 4279 real-world underwater images with more abundant underwater scenes (water types, lighting conditions and target categories) than existing underwater datasets \cite{Liu2020RealWorldUE,Li2019UIEB,Akkaynak2019seathru,li2017watergan}, and the corresponding clear images are generated as comparison references. We also provide the semantic  segmentation  map and medium  transmission  map for each raw underwater image.  The top of each image group is the clear reference image, followed by the raw underwater image, semantic segmentation map, and medium transmission map.}
	\label{figure:LSUI}
\end{figure*}

\noindent
\textbf{Data Collection.} 
We have collected 8018 underwater images, which is composed of images collected by ourself and from other existing public datasets \cite{Liu2020RealWorldUE,Akkaynak2019seathru,li2017watergan,Fabbri2018UGAN} (All images have been licensed and used only for academic purposes). Real underwater images with rich water scenes, water types, lighting conditions and target categories, are selected to the extent possible, for further generating clear reference images.

\noindent
\textbf{Reference Image Generation.} 
The reference images were selected with two round subjective and objective evaluations, to eliminating the potential bias to the extent possible. In the first round, inspired by ensemble learning \cite{polikar2012ensemble} that multiple weak classifiers could form a strong one, we firstly use 18 existing optimal UIE methods \cite{Ancuti2012fusion,Fu2014Retinexbased,peng2017underwater,Drews2013UDCP,Drews2016RBSI,Li2016miniinfor,Chiang2011WCD,galdran2015automatic,Li2016miniop,Islam2020FUnIE,li2017watergan,li2018cyclegan,Yang2011LDCP,Fu2017twostep,song2018rapid,uplavikar2019UIEDAL,qi2022sguienet,ma2022wavelet} to process the collected underwater images successively, and a set with $18 * 8018$ images is generated for the next-step optimal reference dataset selection. Unlike \cite{Li2019UIEB}, to reducing the number of images that need to be selected manually, non-reference metrics UIQM \cite{Panetta2015UIQM} and UCIQE \cite{Yang2015UCIQE} are adopted to score all generated images with equal weights. Then, the top-three reference images of each original one form a set with the size $3 * 8018$. Considering individual differences, 20 volunteers with image processing experience were invited to rate images according to 5 most important judgments (contrast; saturation; color correction effects; artifacts degree; over or under-enhancement degree) of UIE tasks with a score from 0-10, where the higher score represents the more contentedness. And the total score of each reference picture is 100 ($5*20$) after normalizing each score to 0-1. The top-one reference image of each raw underwater image was chosen with the highest summation value. In addition, images with the highest summation lower than 70 have been removed from the dataset.


After the first round, some of the generated reference images still have problems such as blur, color cast and noise. So in the second round, we invited volunteers to vote on each reference picture again to select its existing problems and determine the corresponding optimization method, and then use appropriate image enhancement methods \cite{zamir2021restormer,Liang2021ICCV,ye2021perceiving} to process it. Next, all volunteers were invited to conduct another round of voting to remove image pairs that more than half of the volunteers were dissatisfied with.
To improve the utility of the LSUI dataset, we also hand-labeled a segmentation map and generated a medium transmission map for each image. Eventually, our LSUI dataset contains 4279 images and the corresponding high-quality reference images, semantic segmentation maps, and medium transmission maps for each image.

As shown in Fig .\ref{figure:Statistics}, compared with UIEB \cite{Li2019UIEB}, our LSUI dataset contains large number of images with richer underwater scenes and object categories. In particular, our LSUI dataset includes deep-sea scenes and underwater cave scenes that are not available in previous underwater datasets. We provide some examples of our LSUI dataset in Fig .\ref{figure:LSUI}, which includes varied underwawter scenes, water types, lighting conditions and target categories. As to the authors' best knowledge, 
LSUI is the largest real underwater image dataset with high quality reference images at the present time, which could facilitate the further development of the UIE methods.



\subsection{U-shape Transformer}
\subsubsection{Overall Architecture} 
\begin{figure*}[ht]
	\centering
	\includegraphics[width=1\linewidth]{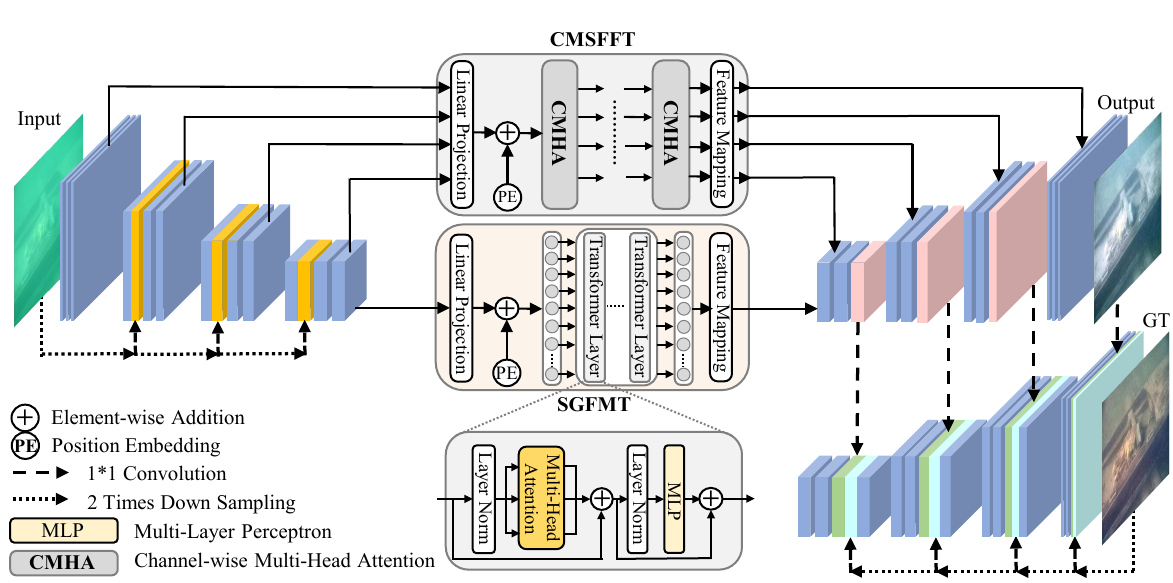}
	\caption{The network structure of U-shape Transformer. CMSFFT and SGFMT modules specially designed for UIE tasks reinforce the network's attention to the more severely attenuated color channels and spatial regions. The multi-scale connections of the generator and the discriminator make the gradient flow freely between the generator and the discriminator, therefore making the training process more stable.}
	\label{figure:network}
\end{figure*}

The overall architecture of the U-shape Transformer is shown as Fig. \ref{figure:network}, which includes a CMSFFT \& SGFMT based generator and a  discriminator. 

In the generator, (1) Encoding: Except being directly input to the network, the original image will be downsampled three times respectively.  Then after 1*1 convolution, the three scale feature maps are input into the corresponding scale convolution block. The outputs of four convolutional blocks are the inputs of the CMSFFT and SGFMT; (2) Decoding: After feature remapping, the SGFMT output is directly sent to the first convolutional block. Meanwhile, four convolutional blocks with varied scales will receive the four outputs from CMSFFT. 

In the discriminator, the input of the four convolutional blocks includes: the feature map output by its own upper layer, the feature map of the corresponding size from the decoding part and the feature map generated by $1 * 1$ convolution after downsampling to the corresponding size using the reference image. 
With the described multi-scale connections, the gradient flow can flow freely on multiple scales between the generator and the discriminator, such that a stable training process could be obtained, details of the generated images could be enriched. The detailed structure of SGFMT and CMSFFT in the network will be described in the following two subsections.

\subsubsection{SGFMT}
\begin{figure}[h] 
	\centering
	\includegraphics[width=1\linewidth]{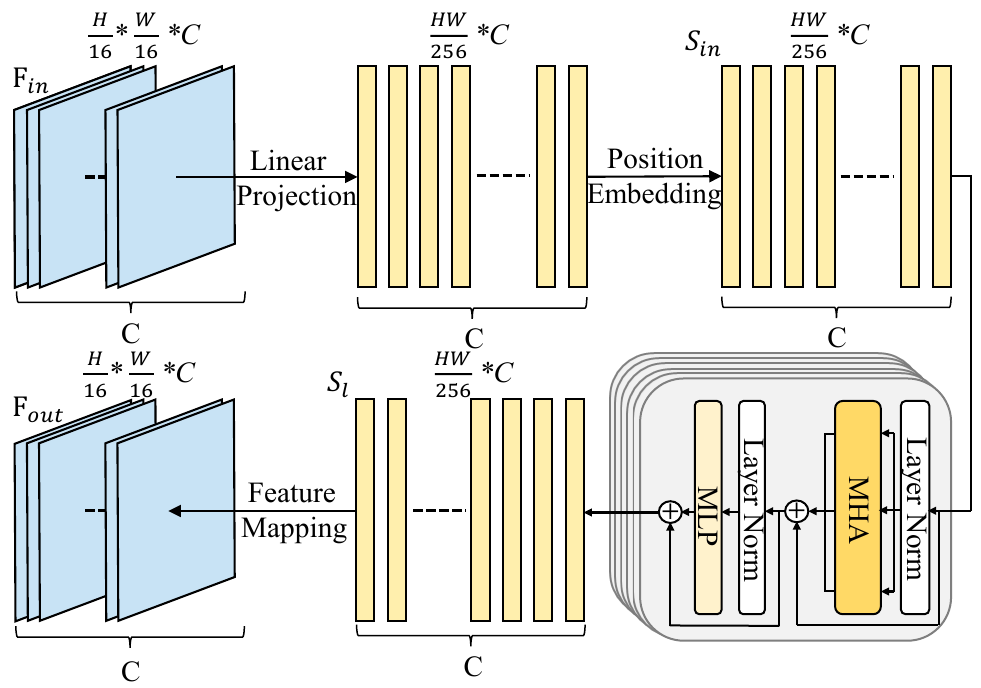}
	\caption{Data flow diagram of the SGFMT module. Based on the prior that underwater images are not uniformly degraded in different spatial regions, we designed a novel \textbf{spatial-wise global feature modeling transformer (SGFMT)} based on the spatial self-attention mechanism to replace the original bottleneck layer of the generator. It can accurately model the global feature of underwater images and reinforce the network's attention to the space areas with more serious attenuation, thus achieving uniform UIE.}
	\label{figure:SGFMT}
\end{figure}
The SGFMT (as shown in Fig. \ref{figure:SGFMT}) is used to replace the original bottleneck layer of the generator, which can assist the network to model the global information and reinforce the network's attention on severely degraded parts. Assuming the size of the input feature map is $F_{in}\in \mathbb{R}^{\frac{H}{16}* \frac{W}{16}\ *C}$.
For the expected one-dimensional sequence of the transformer, linear projection is used to stretch the two-dimensional feature map into a feature sequence $S_{in}\in \mathbb{R}^{\frac{HW}{256}\ast\ C}$. 
For preserving the valued position information of each region, learnable position embedding is merged directly, which can be expressed as,
\begin{equation}
	S_{in}=W\ast F_{in}+{\rm{PE}},
	\label{eq:1}
\end{equation}
where $W*F_{i}$ represents a linear projection operation, PE represents a position embedding operation. 

Then we input the feature sequence $S_{in}$ to the transformer block, which contains  4 standard transformer layers \cite{vaswani2017attention}.
Each transformer layer contains a multi-head attention block (MHA) and a feed-forward network (FFN). The FFN includes a normalization layer and a fully connected layer. 
The output of the $l$-th$(l\in[1,2,\ldots.,l])$ layer in the transformer block can be calculated by,
\begin{equation}
	S_{l}^{'}={\rm MHA}({\rm LN}(S_{l-1}))+S_{l-1}
	\label{eq:2}
\end{equation}
\begin{equation}
	S_{l}={\rm FFN}({\rm LN}(S_{l}^{'}))+S_{l}^{'},
	\label{eq:3}
\end{equation}
where LN represents layer normalization, and $S_l$ represents the output sequence of the $l$-th layer in the transformer block. The output feature sequence of the last transformer block is $S_l\in \mathbb{R}^{\frac{HW}{256}\ast\ C}$, which is restored to the feature map of  $F_{out}\in \mathbb{R}^{\frac{H}{16}\ast\ \frac{W}{16}\ \ast C}$ after feature remapping.


\subsubsection{CMSFFT}

\begin{figure*}[htb]
	\centering
	\includegraphics[width=1\linewidth]{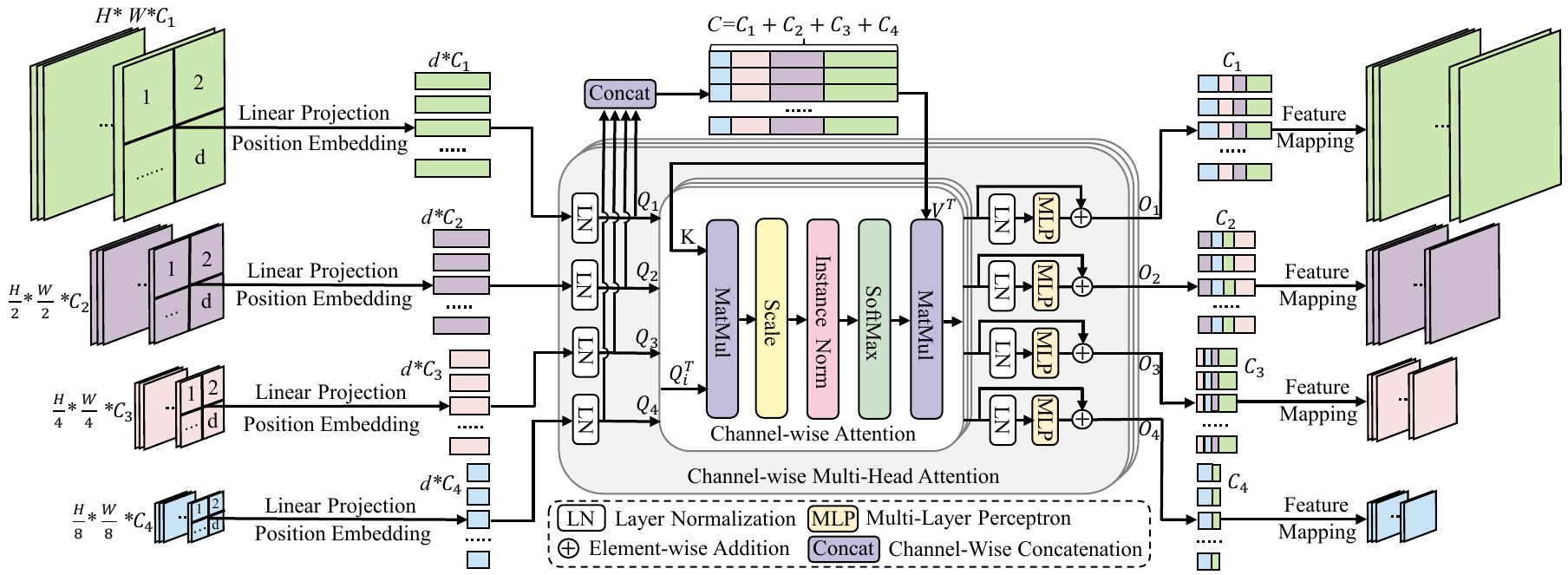}
	\caption{Detailed structure of the CMSFFT module. According to the prior that underwater images are inconsistently attenuated in different color channels, we propose a novel \textbf{channel-wise multi-scale feature fusion transformer(CMSFFT)}. Specifically, CMSFFT replaces the skip connection of the generator, uses the channel-wise self-attention mechanism to perform channel-wise multi-scale feature fusion on the features output by the encoder of the generator, and transmits the fusion results to the decoder efficiently, so as to reinforce the network's attention to the color channels with more serious attenuation and realize accurate UIE.}
	\label{figure:CMSFFT}
\end{figure*}


To reinforce the network's attention on the more serious attenuation color channels, inspired by \cite{wang2021uctransnet}, we designed the CMSFFT block to replace the skip connection of the original generator's encoding-decoding architecture (Fig.\ref{figure:CMSFFT}), which consists of the following three parts.

\noindent
\textbf{Multi-Scale Feature Encoding.}
The inputs of CMSFFT are the feature maps $F_i\in \mathbb{R}^{\frac{H}{2^i} *\frac{W}{2^i} *C_i}(i=0,1,2,3)$ with different scales. Differs from the linear projection in Vit \cite{Dosovitskiy2021AnII} which is applied directly on the partitioned original image, we use convolution kernels with related filter size $\frac{P}{2^i} *\frac{P}{2^i} (i=0,1,2,3)$ and step size $\frac{P}{2^i} (i=0,1,2,3)$, to conduct linear projection on feature maps with varied scales. In this work, $P$ is set as 32. After that, four feature sequence $S_i\in \mathbb{R}^{d\ast\ C_i}(i=1,2,3,4)$ could be obtained, where $d\in \frac{HW}{P^2}$. Those four convolution kernels divide feature maps into the same number of blocks, while the number of channels ${C_i}(i=1,2,3,4)$ remains unchanged. Then, four query vectors $Q_i\in \mathbb{R}^{d\ast\ C_i}(i=1,2,3,4) $, $K\in \mathbb{R}^{d\ast\ C}$ and $V\in \mathbb{R}^{d\ast\ C}$ can be obtained by Eq.(\ref{eq:4}).
\begin{equation}
	Q_i=S_i W_{Q_i} \quad
	K= S W_K \quad
	V=S W_V,
	\label{eq:4}
\end{equation}
where $W_{Q_i}\in \mathbb{R}^{d\ast\ C_i}(i=1,2,3,4) $, $W_K\in \mathbb{R}^{d\ast\ C}$ and $W_V\in \mathbb{R}^{d\ast\ C}$ stands for learnable weight matrices; S is generated by concatenating $S_i\in \mathbb{R}^{d\ast\ C_i}(i=1,2,3,4) $ via the channel dimension, where $C=C_1+C_2+C_3+C_4$. In this work, $C_1$, $C_2$, $C_3$, and $C_4$ are set as 64, 128, 256, 512, respectively.

\noindent
\textbf{Channel-Wise Multi-Head Attention(CMHA).} 
The CMHA block has six inputs, which are $K\in \mathbb{R}^{d\ast\ C}$, $V\in \mathbb{R}^{d\ast\ C}$ and $Q_i\in \mathbb{R}^{d\ast\ C_i}(i=1,2,3,4)$.
The output of channel-wise attention ${\rm CA_i}\in \mathbb{R}^{C_i\ast\ d}(i=1,2,3,4)$ could be obtained by,
\begin{equation}
	\begin{aligned}
		{\rm CA}_{i}&={\rm SoftMax}({\rm IN}(\frac  {Q_{i}^{T} K }{\sqrt[2]{C}})) V^{T},
		\label{eq:5}
	\end{aligned}
\end{equation}
where IN represents the instance normalization operation. This attention operation performs along the channel-axis instead of the classical patch-axis\cite{Dosovitskiy2021AnII}, which can guide the network to pay attention to channels with more severe image quality degradation. 
In addition, IN is used on the similarity maps to assist the gradient flow spreads smoothly. 

The output of the $i$-th CMHA layer can be expressed as,
\begin{equation}
	{\rm CMHA_{i}}=({\rm{CA}_{i}}^{1}+{\rm{CA}_{i}}^{2}+.......,+{\rm{CA}_{i}}^{N})/N+Q_{i},
	\label{eq:6}
\end{equation}
where $N$ is the number of heads, which is set as 4 in our implementation.

\noindent
\textbf{Feed-Forward Network(FFN).}
Similar to the forward propagation of  \cite{Dosovitskiy2021AnII}, the FFN output can be expressed as, 
\begin{equation}
	O_{i}=\rm{CMHA}_{i}+\rm{MLP}(\rm{LN}(\rm{CMHA_{i}})),
	\label{eq:7}
\end{equation}
where $O_i\in \mathbb{R}^{d\ast\ C_i}(i=1,2,3,4)$; MLP stands for multi-layer perception. Here, The operation in Eq. (\ref{eq:7}) needs to be repeated $l$ ($l$=4 in this work) times in sequence to build the $l$-layer transformer.


Finally, feature remappings are performed on the four different output feature sequences $O_i\in \mathbb{R}^{C_i\ast\ d}(i=1,2,3,4)$ to reorganize them into four feature maps $F_i\in \mathbb{R}^{\frac{H}{2^i} *\frac{W}{2^i} *C_i}(i=0,1,2,3)$ , which are the input of convolutional block in the generator's decoding part.

\begin{table*}[b]
	\centering
	\caption{Statistical results of color space selection experiments. We test U-shape Transformers trained with different color space loss functions on Test-L400 and Test-U90 datasets, respectively, and the color spaces that obtain the top three PSNR scores are marked with red, green, and blue, respectively.}
	\begin{tabular}{c||cccccccc}
		\hline
		\begin{tabular}[c]{@{}l@{}}Color\\ Space\end{tabular} & RGB   & HSV   & HSI   & XYZ   & LAB   & LUV   & LCH   & YUV   \\ \hline \hline
		Tset-L400                                             & \textcolor{green}{\textbf{23.79}} & 23.32 & 23.37 & 22.63 & \textcolor{red}{\textbf{23.86}} & 22.81 & \textcolor{blue}{\textbf{23.62}} & 23.43 \\ 
		Test-U90                                              & \textcolor{red}{\textbf{22.72}} & 22.01 & 22.17 & 21.69 & \textcolor{green}{\textbf{22.53}} & 21.77 & \textcolor{blue}{\textbf{22.49}} & 22.23 \\ \hline
	\end{tabular}
	\label{Tab:1}
\end{table*}
\subsection{Loss Function}
To take advantage of the LAB and LCH color spaces' wider color gamut representation range and more accurate description of the color saturation and brightness,
we designed a multi-color space loss function combining RGB, LAB and LCH color spaces to train our network. 
The image from RGB space is firstly converted to LAB and LCH space, and reads,
\begin{equation}
\begin{aligned}
	&L^{G(x)},A^{G(x)},B^{G(x)}=\rm{RGB2LAB}(G(x))\\
	&L^{y},A^{y},B^{y}=\rm{RGB2LAB}(y),
	\label{eq:8}
\end{aligned}
\end{equation}
\begin{equation}
\begin{aligned}
	&L^{G(x)},C^{G(x)},H^{G(x)}=\rm{RGB2LCH}(G(x)),\\
	&L^{y},C^{y},H^{y}=\rm{RGB2LCH}(y),
	\label{eq:9}
\end{aligned}
\end{equation}
where $x$, $y$ and $G(x)$ represents the original inputs, the reference image, and the clear image output by the generator, respectively.

Loss functions in the LAB and LCH space are written as Eq.(\ref{eq:10}) and Eq.(\ref{eq:11}).
\begin{equation}
	\begin{aligned}
		&Loss_{LAB}(G(x),y)=E_{x,y}[{(L^y-L^{G(x)})}^2-\\
		&\sum_{i=1}^{n}{Q(A_i^y)log(Q(A_i^{G(x)}))-\sum_{i=1}^{n}{Q(B_i^y)log(Q(B_i^{G(x)}))}}],
		\label{eq:10}	 
	\end{aligned}
\end{equation}
\begin{equation}
	\begin{aligned}
		&Loss_{LCH}(G(x),y)=E_{x,y}[-\sum_{i=1}^{n}{Q(L_i^y)log(Q(L_i^{G(x)}))}\\
		&+{(C^y-C^{G(x)})}^2+{(H^y-H^{G(x)})}^2],
		\label{eq:11}	 
	\end{aligned}
\end{equation}
where $Q$ stands for the quantization operator. 

${L}_2$ loss in the RGB color space $Loss_{RGB}$ and the perceptual loss $Loss_{per}$\cite{Johnson2016Perceptual}
, as well as $Loss_{LAB}$ and $Loss_{LCH}$ are the four loss functions for the generator.

Besides, standard GAN loss function is introduced for minimizing the loss between generated and reference pictures, and written as,
\begin{equation}
	\begin{aligned}
		L_{GAN}(G,D)=E_y[logD(y)]+E_x[log(1-D(G(x)))],
		\label{eq:14}	 
	\end{aligned}
\end{equation}
where $D$ represents the discriminator. D aims at maximizing $L_{GAN}(G,D)$, to accurately distinguish the generated image from the reference image. And the goal of generator G is to minimize the loss between generated pictures and reference pictures.

Then, the final loss function is expressed as,
\begin{equation}
	\begin{aligned}
		&G^{*}=arg\mathop{\min}\limits_{G}\mathop{\max}\limits_{D}L_{GAN}(G,D)+\alpha{Loss}_{LAB}(G(x),y)\\
		&+\beta{Loss}_{LCH}(G(x),y)+\gamma{Loss}_{RGB}(G(x),y)\\
		&+\mu{Loss}_{per}(G(x),y),
		\label{eq:15}
	\end{aligned}
\end{equation}
where $\alpha, \beta, \gamma, \mu$ are hyperparameters, which are set as 0.001, 1, 0.1, 100, respectively, with numerous experiments.

\section{Experiments}
In this section,  we first introduce the training details of the U-shape Transformer and the detailed settings of the experiment. Next, we conduct experiments on the selection of color space. Then we retrain some network models we collected on the existing underwater datasets and the LSUI dataset to evaluate our proposed dataset. Moerover, we also compare our UIE method with state-of-the-arts on five datasets.  Finally, series of ablation studies are conducted to demonstrate the effectiveness of each component in U-shape Transformer.

\subsection{Implementation Details}
The LSUI dataset was randomly divided as Train-L (3879 images) and Test-L400 (400 images) for training and testing, respectively. The training set was enhanced by cropping, rotating and flipping the existing images. All images were adjusted to a fixed size (256*256) when input to the network, and the pixel value will be normalized to [0,1].

We use python and pytorch framework via NVIDIA RTX3090 on Ubuntu20 to implement the U-shape Transformer. Adam optimization algorithm is utilized for the total of 800 epochs training with batchsize set as 6. The initial learning rate is set as 0.0005 and 0.0002 for the first 600 epochs and the last 200 epochs, respectively. Besides, the learning rate decreased 20\% every 40 epochs. For $ Loss_{RGB}$, $L_2$ loss is used for the first 600 epochs, and $L_1$ loss is used for the last 200 epochs.

\subsection{Experiment Settings}
\noindent
\textbf{Benchmarks.}
Besides Train-L, the second training set Train-U contains 800 pairs of underwater images from UIEB \cite{Li2019UIEB} and 1,250 synthetic underwater images from \cite{li2020underwater}; the third training set Train-E contains the paired training images in the EUVP \cite{Islam2020FUnIE} dataset. Testing datasets are categorized into two types, (1) full-reference testing dataset: Test-L400 and Test-U90 (remaining 90 pairs in UIEB); (2) non-reference testing dataset: Test-U60 and SQUID. Here, Test-U60 includes 60 non-reference images in UIEB; 16 pictures from SQUID \cite{Akkaynak2019seathru} forms the second non-reference testing dataset.

\noindent
\textbf{Compared Methods.}
We compare U-shape Transformer with 10 UIE methods to verify our performance superiority. It includes two physical-based models (UIBLA \cite{peng2017underwater}, UDCP \cite{Drews2013UDCP}), three visual prior-based methods (Fusion \cite{Ancuti2012fusion}, retinex based \cite{Fu2014Retinexbased}, RGHS \cite{huang2018RGHS}), and five data-driven methods (WaterNet \cite{Li2019UIEB}, FUnIE \cite{Islam2020FUnIE}, UGAN \cite{Fabbri2018UGAN}, UIE-DAL \cite{uplavikar2019UIEDAL}, Ucolor \cite{Li2021UnderwaterIE}). 

\begin{figure*}[h]
	\centering
	\includegraphics[width=1\linewidth]{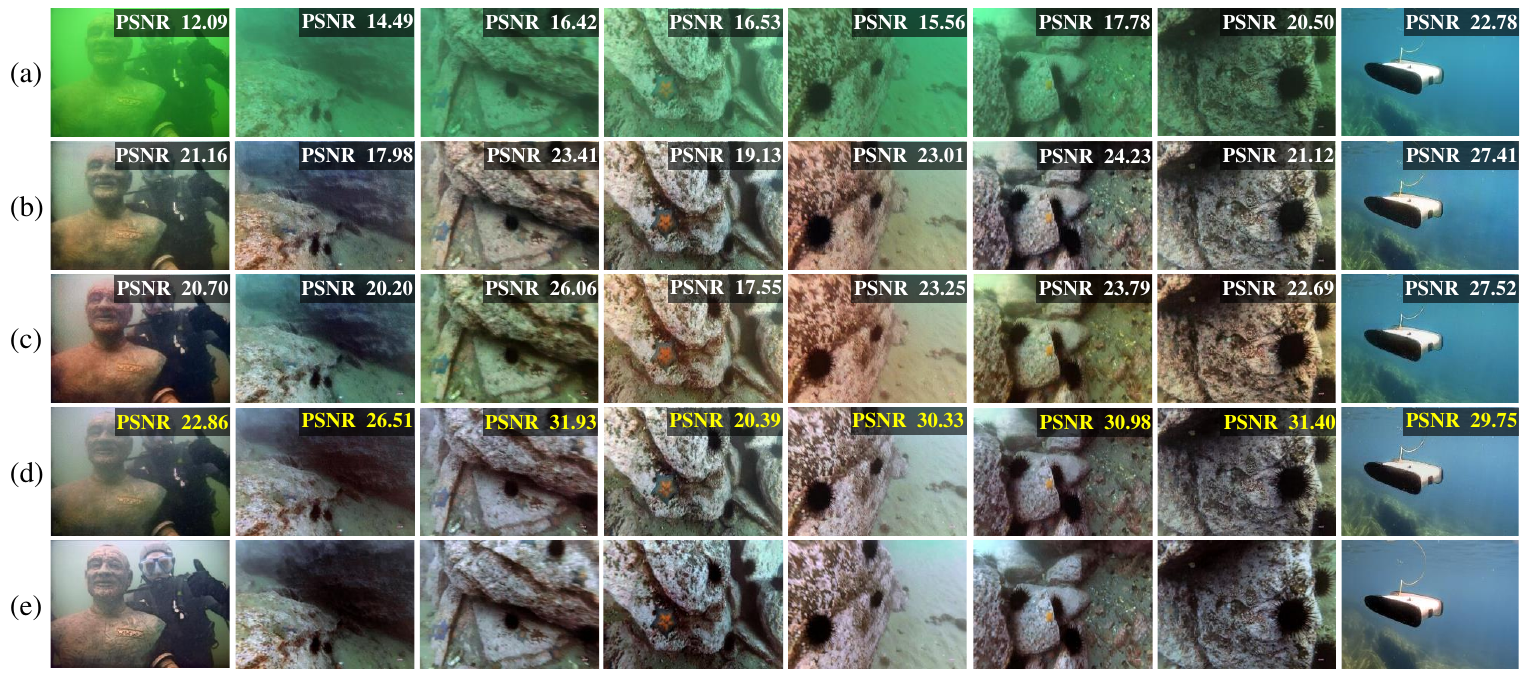}
	\caption{Enhancement results of U-shape transformer trained on different underwater datasets. (a): Input images; (b): Enhanced results using the model trained on the Train-U; (c): Enhanced results using the model
		trained on the Train-E; (d): Enhanced results using the model trained by our proposed dataset Train-L; (e): Reference images(recognized as ground truth (GT)).}
	\label{figure:dataset_evaluation}
\end{figure*}
\noindent
\textbf{Evaluation Metrics.}
For the testing dataset with reference images, we conducted full-reference evaluations using PSNR \cite{korhonen2012peak} and SSIM \cite{Alain2010SSIM} metrics. Those two metrics reflect the proximity to the reference, where a higher PSNR value represents closer image content, and a higher SSIM value reflects a more similar structure and texture.
For images in the non-reference testing dataset, non-reference evaluation metrics UCIQE \cite{Yang2015UCIQE} and UIQM \cite{Panetta2015UIQM} are employed, in which higher UCIQE or UIQM score suggests better human visual perception. 
For UCIQE and UIQM cannot accurately measure the performance in some cases \cite{Li2019UIEB} \cite{Berman2020HLQD}, we also conducted a survey following \cite{Li2021UnderwaterIE}, which results are stated as “perception score (PS)". PS ranges from 1-5, with higher scores indicating higher image quality.
Moreover, NIQE \cite{Mittal2012NIQE}, which lower value represents a higher visual quality, is also adopted as the metrics.

\subsection{Color Space Selection}
\label{Sec:E}

In order to select the appropriate color space to form the multi-color space loss function, we use the mixed loss function composed of the single color space loss function and other loss functions to train the U-shape transformer. We use Train-L to train the network, and then test and calculate PSNR on Test-L400 and Test-U90 data sets, respectively. The results are shown in Tab. \ref{Tab:1},

As in Tab. \ref{Tab:1}, We note that the LAB, LCH, and RGB color spaces achieve the top-3 PSNR scores on both test datasets. In RGB color space, image is easy to store and display because of its strong color physical meaning, but these three components (R, G, and B) are highly correlated and easily affected by brightness, shadows, noise, and other factors. Compared with other color spaces, LAB color space is more consistent with the characteristics of human visual, can express all colors that human eyes can perceive, and the color distribution is more uniform. LCH color space can intuitively express brightness, saturation, and hue. Combined with the experimental results and the above analysis, we choose LAB, LCH, and RGB color space to form our multi-color space loss function.

\subsection{Dataset Evaluation}
\label{section:43}

\begin{table}[h]
	\centering
	\caption{Dataset evaluation results. The highest PSNR and SSIM scores are marked in red.}
	\begin{tabular}{c||c|cc|cc}
		\hline
		\multirow{2}{*}{Methods}             & \multirow{2}{*}{\begin{tabular}[c]{@{}c@{}}Training\\  Data\end{tabular}} & \multicolumn{2}{c|}{Test-U90}     & \multicolumn{2}{c}{Test-L400}     \\ \cline{3-6} 
		&                                & \multicolumn{1}{l|}{PSNR}  & SSIM & \multicolumn{1}{l|}{PSNR}  & SSIM \\ \hline\hline
		\multirow{3}{*}{U-net\cite{ronneberger2015u}}               & Train-U                   & \multicolumn{1}{c|}{17.07} & 0.76 & \multicolumn{1}{c|}{19.19} & 0.79 \\  
		& Train-E                   & \multicolumn{1}{l|}{17.46} & 0.76 & \multicolumn{1}{l|}{19.45} & 0.78 \\  
		& Ours                     & \multicolumn{1}{l|}{\textcolor{red}{\textbf{20.14}}} & \textcolor{red}{\textbf{0.81}} & \multicolumn{1}{l|}{\textcolor{red}{\textbf{20.89}}} & \textcolor{red}{\textbf{0.82}} \\ \hline
		\multirow{3}{*}{UGAN\cite{Fabbri2018UGAN}}                & Train-U                  & \multicolumn{1}{c|}{20.71} & 0.82 & \multicolumn{1}{c|}{19.89} & 0.79 \\  
		& Train-E                  & \multicolumn{1}{c|}{20.72} & 0.82 & \multicolumn{1}{c|}{19.82} & 0.78 \\ 
		& Ours                     & \multicolumn{1}{c|}{\textcolor{red}{\textbf{21.56}}} & \textcolor{red}{\textbf{0.83}} & \multicolumn{1}{c|}{\textcolor{red}{\textbf{21.74}}} & \textcolor{red}{\textbf{0.84}} \\ \hline
		\multirow{3}{*}{Ours} & Train-U                   & \multicolumn{1}{c|}{21.25} & 0.84 & \multicolumn{1}{c|}{22.87} & 0.85 \\ 
		& Train-E                   & \multicolumn{1}{c|}{21.75} & 0.86 & \multicolumn{1}{c|}{23.01} & 0.87 \\ 
		& Ours                     & \multicolumn{1}{c|}{\textcolor{red}{\textbf{22.91}}} & \textcolor{red}{\textbf{0.91}} & \multicolumn{1}{c|}{\textcolor{red}{\textbf{24.16}}} & \textcolor{red}{\textbf{0.93}} \\ \hline
	\end{tabular}
	\label{Tab:2}
\end{table}

The effectiveness of LSUI is evaluated by retraining the compared methods (U-net \cite{ronneberger2015u}, UGAN \cite{Fabbri2018UGAN} and U-shape Transformer) on Train-L, Train-U and Train-E. The trained network was tested on Test-L400 and Test-U90. 

As shown in Tab.\ref{Tab:2}, the model trained on our dataset is the best of PSNR and SSIM. It could be explained that LSUI contains richer underwater scenes and better visual quality reference images than existing underwater image datasets, which could improve the enhancement and generalization ability of the tested network.

Fig. \ref{figure:dataset_evaluation} is the sampled enhancement results of U-shape transformer trained on different underwater datasets, which is a supplement of the Data Evaluation part of the paper. Enhancement results training on Train-L (a portion of our LSUI dataset) demonstrates the highest PSNR value and preferable visual quality, while results training on other datasets show a certain degree of color cast. 
For the high-quality reference images and rich underwater scenes (lighting conditions, water types and target categories), our constructed LSUI dataset could improve the imaging quality and generalization performance of the UIE network.



\begin{table*}[h]
	\centering
	\caption{Quantitative comparison among different UIE methods on the full-reference testing set. The highest scores of PSNR and SSIM are marked in red, and all UIE methods are tested on a PC with an INTEL(R) I5-10500 CPU, 16.0GB RAM, a NVIDIA GEFORCE RTX 1660 SUPER.}
	\begin{tabular}{c||cc|cc|c|c|c}
		\hline
		\multirow{2}{*}{Methods} & \multicolumn{2}{c|}{Test-L400}     & \multicolumn{2}{c|}{Test-U90}      & \multirow{2}{*}{FLOPs↓} & \multirow{2}{*}{\#param.↓} & \multirow{2}{*}{time↓} \\ \cline{2-5}
		& \multicolumn{1}{c|}{PSNR↑} & SSIM↑ & \multicolumn{1}{c|}{PSNR↑} & SSIM↑ &                            &                                 &                                   \\ \hline\hline
		UIBLA\cite{peng2017underwater} & \multicolumn{1}{c|}{13.54} & 0.71  & \multicolumn{1}{c|}{15.78} & 0.73  & $\times$                   &$\times$                         & 42.13s                            \\ 
		UDCP\cite{Drews2013UDCP}       & \multicolumn{1}{c|}{11.89} & 0.59  & \multicolumn{1}{c|}{13.81} & 0.69  &$\times$                    & $\times$                        & 30.82s                            \\ 
		Fusion\cite{Ancuti2012fusion}  & \multicolumn{1}{c|}{17.48} & 0.79  & \multicolumn{1}{c|}{19.04} & 0.82  & $\times$                   & $\times$                        & 6.58s                        \\ 
		Retinex based\cite{Fu2014Retinexbased} & \multicolumn{1}{c|}{13.89} & 0.74  & \multicolumn{1}{c|}{14.01} & 0.72  & $\times$                   &  $\times$                       & 1.06s                                  \\ 
		RGHS\cite{huang2018RGHS} & \multicolumn{1}{c|}{14.21} & 0.78  & \multicolumn{1}{c|}{14.57} & 0.79  &$\times$                    & $\times$                        &8.92s                                   \\ 
		WaterNet\cite{Li2019UIEB}& \multicolumn{1}{c|}{17.73} & 0.82  & \multicolumn{1}{c|}{19.81} & 0.86  & 193.7G                      & 24.81M                           & 0.61s                             \\ 
		FUnIE\cite{Islam2020FUnIE} & \multicolumn{1}{c|}{19.37} & 0.84  & \multicolumn{1}{c|}{19.45} & 0.85  & 10.23G                      & 7.019M                           & 0.09s                             \\ 
		UGAN\cite{Fabbri2018UGAN} & \multicolumn{1}{c|}{19.79} & 0.78  & \multicolumn{1}{c|}{20.68} & 0.84  & 38.97G                      & 57.17M                           & 0.05s                             \\ 
		UIE-DAL\cite{uplavikar2019UIEDAL}  & \multicolumn{1}{c|}{17.45} & 0.79  & \multicolumn{1}{c|}{16.37} & 0.78  & 29.32G                      & 18.82M                           & 0.07s                             \\ 
		Ucolor\cite{Li2021UnderwaterIE}  & \multicolumn{1}{c|}{22.91} & 0.89  & \multicolumn{1}{c|}{20.78} & 0.87  & 443.85G                     & 157.4M                           & 2.75s                             \\ 
		Ours                     & \multicolumn{1}{c|}{\textcolor{red}{\textbf{24.16}}} & \textcolor{red}{\textbf{0.93}}  & \multicolumn{1}{c|}{\textcolor{red}{\textbf{22.91}}} & \textcolor{red}{\textbf{0.91}}  & 66.2G  & 65.6M & 0.07s         \\ \hline
	\end{tabular}
	\label{Tab:3}
\end{table*}

\begin{table*}[h]
	\centering
	\caption{Quantitative comparison among different UIE methods on the non-reference testing set. The highest scores are marked in red. }
	\begin{tabular}{c||c|c|c|c|c|c|c|c|c}
		\hline
		\multicolumn{2}{c||}{\multirow{2}{*}{Methods}} & \multicolumn{4}{c|}{Test-U60} & \multicolumn{4}{c}{SQUID}    \\ \cline{3-10} 
		\multicolumn{2}{c||}{}                         & PS↑  & UIQM↑ & UCIQE↑ & NIQE↓ & PS↑  & UIQM↑ & UCIQE↑ & NIQE↓ \\ \hline \hline
		\multicolumn{2}{c||}{input}                    & 1.46 & 0.82  & 0.45   & 7.16  & 1.23 & 0.81  & 0.43   & 4.93  \\ 
		\multicolumn{2}{c||}{UIBLA\cite{peng2017underwater}}                    & 2.18 & 1.21  & 0.60   & 6.13  & 2.45 & 0.96  & 0.52   & 4.43  \\ 
		\multicolumn{2}{c||}{UDCP\cite{Drews2013UDCP}}                     & 2.01 & 1.03  & 0.57   & 5.94  & 2.57 & 1.13  & 0.51   & 4.47  \\ 
		\multicolumn{2}{c||}{Fusion\cite{Ancuti2012fusion}}                   & 2.12 & \textcolor{red}{\textbf{1.23}}  & 0.61   & 4.96  & 2.89 & \textcolor{red}{\textbf{1.29}}  & 0.61   & 5.01  \\ 
		\multicolumn{2}{c||}{Retinex based\cite{Fu2014Retinexbased}}          & 2.04 & 0.94  & 0.69   & 4.95  & 2.33 & 1.01  & 0.66   & 4.86  \\ 
		\multicolumn{2}{c||}{RGHS\cite{huang2018RGHS}}                    & 2.45 & 0.66  & 0.71   & 4.82  & 2.67 & 0.82  & \textcolor{red}{\textbf{0.73}}   & 4.54  \\ 
		\multicolumn{2}{c||}{WaterNet\cite{Li2019UIEB}}                 & 3.23 & 0.92  & 0.51   & 6.03  & 2.72 & 0.98  & 0.51   & 4.75  \\ 
		\multicolumn{2}{c||}{FUnIE\cite{Islam2020FUnIE}}                    & 3.12 & 1.03  & 0.54   & 6.12  & 2.65 & 0.98  & 0.51   & 4.67  \\ 
		\multicolumn{2}{c||}{UGAN\cite{Fabbri2018UGAN}}                     & 3.64 & 0.86  & 0.57   & 6.74  & 2.79 & 0.90  & 0.58   & 4.56  \\ 
		\multicolumn{2}{c||}{UIE-DAL\cite{uplavikar2019UIEDAL}}                  & 2.03 & 0.72  & 0.54   & 4.99  & 2.21 & 0.79  & 0.57   & 4.88  \\ 
		\multicolumn{2}{c||}{Ucolor\cite{Li2021UnderwaterIE}}                   & 3.71 & 0.84  & 0.53   & 6.21  & 2.82 & 0.82  & 0.51   & 4.32  \\ 
		\multicolumn{2}{c||}{Ours}     & \textcolor{red}{\textbf{3.91}} & 0.85  & \textcolor{red}{\textbf{0.73}}   & \textcolor{red}{\textbf{4.74}}  & \textcolor{red}{\textbf{3.23}} & 0.89  & 0.67   & \textcolor{red}{\textbf{4.24}}  \\ \hline
	\end{tabular}
	\label{Tab:4}
\end{table*}

\begin{table*}[h]
	\centering
	\caption{The color dissimilarity comparisons of different methods on color-check7 in terms of the ciede2000. The best scores are marked in red.}
	\begin{tabular}{c||c|c|c|c|c|c|c|c}
		\hline
		Methods       & Pen W60 & Pen W80 & Can D10 & Fuj Z33 & Oly T6000 & Oly T8000 & Pan TS1 & Avg   \\ \hline\hline
		input         & 14.21   & 16.92   & 17.14   & 16.03   & 15.02     & 22.43     & 18.65   & 17.2  \\ 
		UIBLA\cite{peng2017underwater}         & 13.45   & 16.31   & 14.48   & 14.29   & 12.46     & 14.91     & 20.13   & 15,15 \\ 
		UDCP\cite{Drews2013UDCP}          & 15.32   & 24.12   & 16.53   & 13.21   & 12.65     & 16.78     & 12.85   & 15.92 \\ 
		Fusion\cite{Ancuti2012fusion}        & 12.65   & 13.54   & 14.43   & 12.31   & 11.78     &\textcolor{red}{\textbf{10.97}}     & 11.15   & 12.41 \\ 
		Retinex based\cite{Fu2014Retinexbased} & 13.08   & 19.25   & 17.13   & 18.85   & 17.18     & 19.45     & 20.62   & 17.94 \\ 
		RGHS\cite{huang2018RGHS}          & 11.07   & 12.73   & 15.92   & 13.47   & 14.26     & 18.73     & 12.06   & 14.03 \\ 
		WaterNet\cite{Li2019UIEB}      & 12.54   & 19.82   & 15.71   & 12.73   & 17.75     & 21.87     & 18.91   & 17.05 \\ 
		FUnIE\cite{Islam2020FUnIE}         & 12.81   & 11.81   & 12.39   & 12.76   & 12.46     & 16.74     & 19.28   & 14.04 \\ 
		UGAN\cite{Fabbri2018UGAN}          & 20.49   & 21.75   & 22.63   & 26.49   & 21.63     & 22.05     & 20.73   & 22.25 \\ 
		UIE-DAL\cite{uplavikar2019UIEDAL}       & 12.94   & 16.73   & 14.64   & 12.93   & 16.78     & 17.21     & 18.34   & 15.65 \\ 
		Ucolor\cite{Li2021UnderwaterIE}        & 9.12    & 11.14   & 12.43   & 10.02   & 8.31      & 14.18     & 13.41   & 11.23 \\ 
		Ours &\textcolor{red}{\textbf{7.87}} &\textcolor{red}{\textbf{9.70}} &\textcolor{red}{\textbf{9.96}} &\textcolor{red}{\textbf{8.23}} &\textcolor{red}{\textbf{7.71}} &11.14 &\textcolor{red}{\textbf{9.81}}  &\textcolor{red}{\textbf{9.20}}  \\ \hline
	\end{tabular}
	\label{Tab:5}
\end{table*}

\begin{figure*}[t]
	\centering
	\includegraphics[width=1\linewidth]{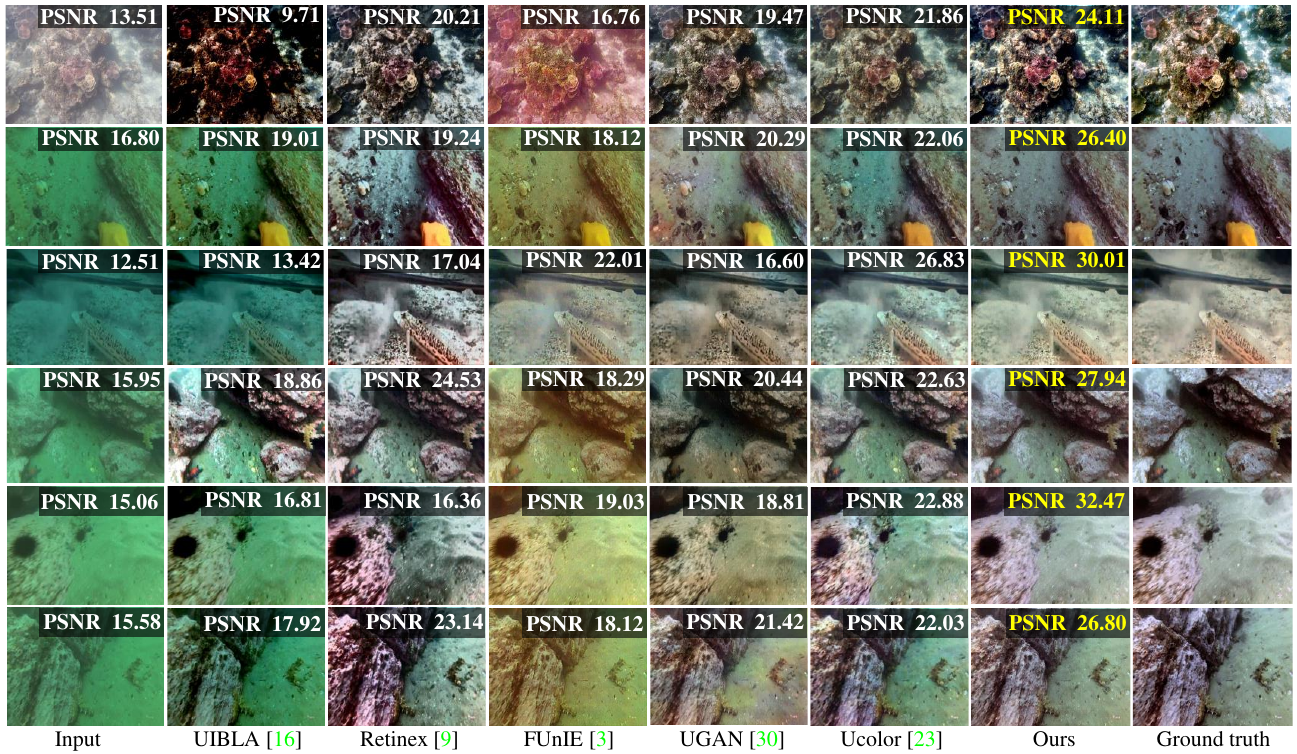}
	\caption{Visual comparison of enhancement results sampled from the Test-L400(LSUI) and Test-U90(UIEB\cite{Li2019UIEB}) dataset. From left to right are raw underwater images, results of UIBLA\cite{peng2017underwater}, Retinex based\cite{Fu2014Retinexbased}, FUnIE\cite{Islam2020FUnIE}, UGAN\cite{Fabbri2018UGAN}, Ucolor\cite{Li2021UnderwaterIE}, our U-shape Transformer and the reference image (recognized as ground truth (GT)). The highest PSNR value of each raw is marked in yellow.}
	\label{figure:full-reference}
\end{figure*}

\subsection{Network Architecture Evaluation}
\label{sec:4.4}

\textbf {Full-Reference Evaluation.} The Test-L400 and Test-U90 datasets were used for evaluation. The statistical results and visual comparisons are summarized in Tab. \ref{Tab:3} and Fig. \ref{figure:full-reference}. We also provide the running time (image size is 256*256) of all UIE methods in Tab. \ref{Tab:3}, as well as the FLOPs and parameter amount of each data-driven UIE method. And we retrianed the 5 open-sourced deep learning-based UIE methods on our dataset.

\noindent
As in Tab.\ref{Tab:3}, our U-shape Transformer demonstrates the best performance on both PSNR and SSIM metrics with relatively few parameters, FLOPs, and running time. The potential limitations of the performance of the 5 data-driven methods are analyzed as follows. The strength of FUnIE \cite{Islam2020FUnIE} lies in achieving fast, lightweight,and fewer parameter models, while naturally limits its scalability on complex and distorted testing samples.
UGAN \cite{Fabbri2018UGAN} and UIE-DAL \cite{uplavikar2019UIEDAL} did not consider the inconsistent characteristics of the underwater images. Ucolor's media transmission map prior can not effectively represent the attenuation of each area, and simply introducing the concept of multi-color space into the network's encoder part cannot effectively take advantage of it, which causes unsatisfactory results in terms of contrast, brightness, and detailed textures. 


The visual comparisons shown in Fig. \ref{figure:full-reference} reveal that enhancement results of our method are the closest to the reference image, which has fewer color artifacts and high-fidelity object areas. Five selected methods tend to produce color artifacts that deviated from the original color of the object. Among the methods,  UIBLA \cite{peng2017underwater} exhibits severe color casts. Retinex based\cite{Fu2014Retinexbased} 
could improve the image contrast to a certain extent, but cannot remove the color casts and color artifacts effectively. The enhancement result of FUnLE \cite{Islam2020FUnIE} is yellowish and reddish overall. Although UGAN \cite{Fabbri2018UGAN} and Ucolor \cite{Li2019UIEB} could provide relatively good color appearance, they are often affected by local over-enhancement, and there are still some color casts in the result.

\begin{figure*}[h]
	\centering
	\includegraphics[width=1\linewidth]{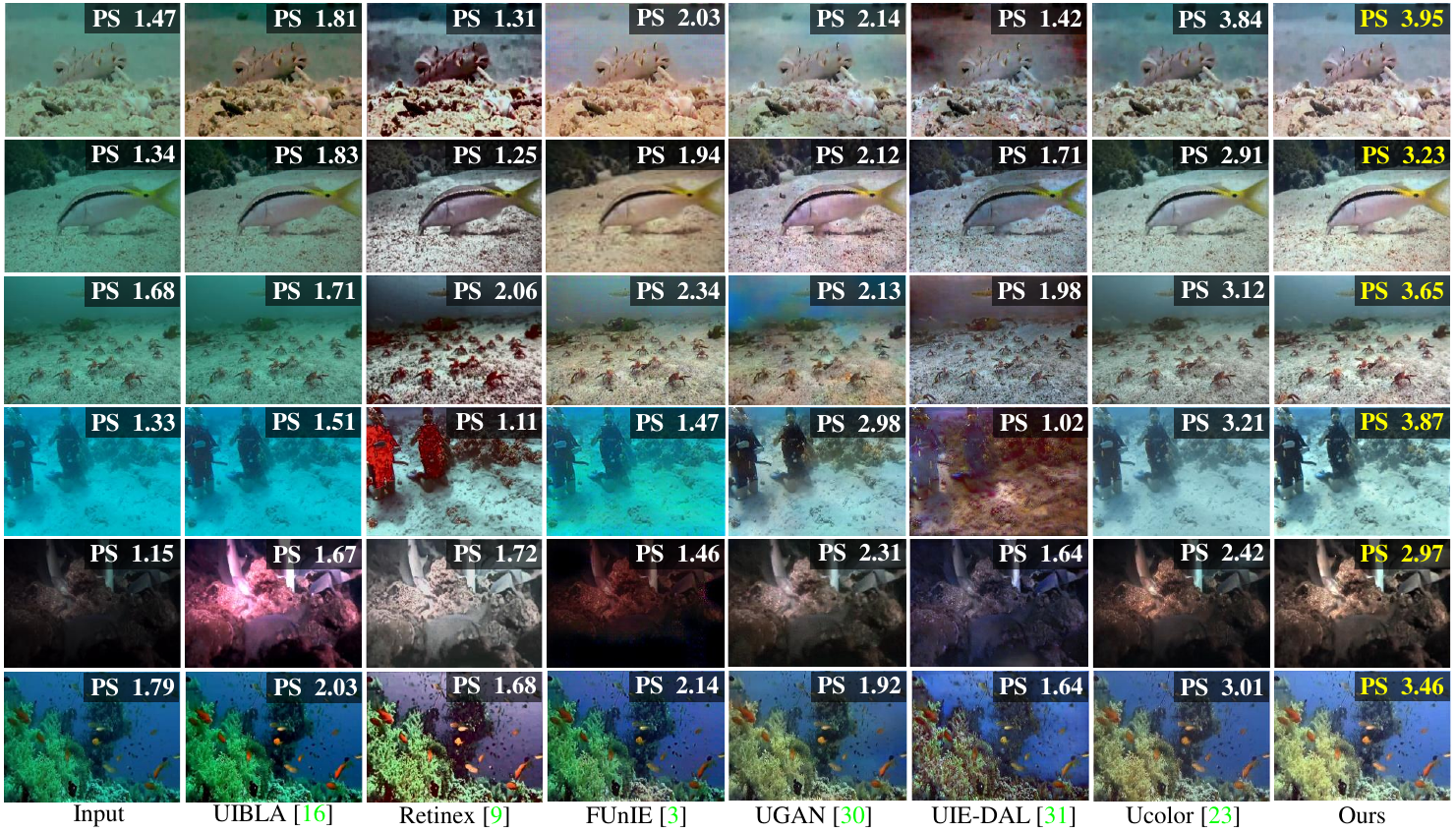}
	\caption{Visual comparison of the non-reference evaluation sampled from the Test-U60(UIEB \cite{Li2019UIEB}) dataset. From left to right are raw underwater images, results of UIBLA \cite{peng2017underwater}, Retinex based \cite{Fu2014Retinexbased}, FUnIE \cite{Islam2020FUnIE}, UGAN \cite{Fabbri2018UGAN}, UIE-DAL \cite{uplavikar2019UIEDAL}, Ucolor \cite{Li2021UnderwaterIE} and our U-shape Transformer. The score in the upper right corner of each image is the perception score(PS), and the highest PS value of each raw is marked in yellow.}
	\label{figure:non-reference}
\end{figure*}

\noindent
\textbf {Non-reference Evaluation.} The Test-U60 and SQUID datasets were utilized for the non-reference evaluation, in which statistical results and visual comparisons are shown in Tab. \ref{Tab:4} and Fig. \ref{figure:non-reference}.

As in Tab. \ref{Tab:4}, our method achieved the highest scores on PS and NIQE metrics, which confirmed the initial idea to contemplate the human eye's color perception and better generalization ability to varied real-world underwater scenes.
Note that UCIQE and UIQM of all deep learning-based UIE methods are weaker than physical model-based or visual prior-based, also reported in \cite{Li2021UnderwaterIE}. 
Those two metrics are of valuable reference, but cannot as absolute justifications \cite{Li2019UIEB}\cite{Berman2020HLQD}, for they are non-sensitive to color artifacts \& casts and biased to some features. 



As in Fig. \ref{figure:non-reference}, enhancement results of our method have the highest PS value, which index reflects the visual quality. Generally, compared methods are unsatisfactory, which includes undesirable color artifacts, over-saturation and unnatural color casts. Among the methods, results of the UIBLA \cite{peng2017underwater} and FUnIE \cite{Islam2020FUnIE} have a certain degree of color cast. Retinex based \cite{Fu2014Retinexbased} method introduces artifacts and unnatural colors. UGAN \cite{Fabbri2018UGAN} and UIE-DAL \cite{uplavikar2019UIEDAL} have the issue of local over-enhancement and color artifacts, which main reason is they ignore the inconsistent attenuation characteristics of the underwater images in the different space areas and the color channels. Although Ucolor \cite{Li2021UnderwaterIE} introduces the transmission medium prior to reinforcing the network's attention on the spatial area with severe attenuation, it still ignores the inconsistent attenuation characteristics of the underwater image in different color channels, which results in the problem of overall color cast. In our method, the reported CMSFFT and SGFMT modules could reinforce the network's attention to the color channels and spatial regions with serious attenuation, therefore obtaining  high visual quality enhancement results without artifacts and color casts.

\begin{figure*}[h]
	\centering
	\includegraphics[width=1\linewidth]{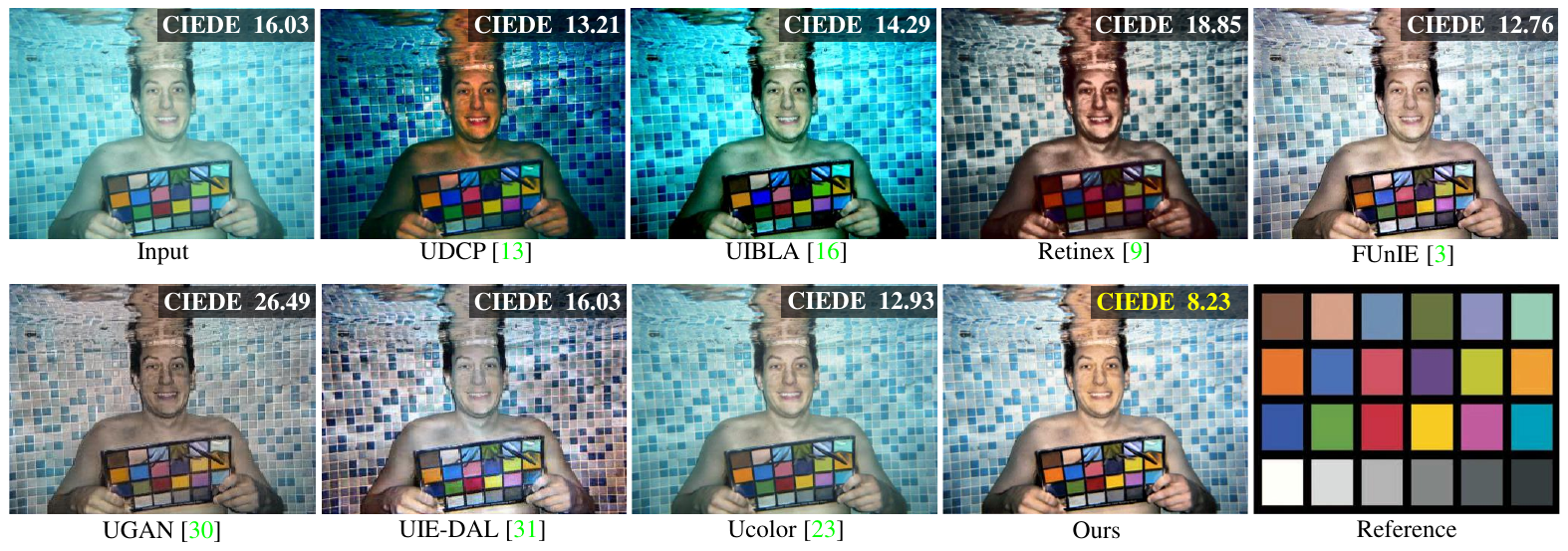}
	\caption{Visual comparison of the color restoration performance evaluation. The input image is sampled from color-check7 dataset and it's taken by Fuji Z33. The values of CIEDE2000 metric for the regions of Color Checker are reported on the top-left corner of the images (the smaller, the better).}
	\label{figure:color_check}
\end{figure*}

\subsection{Color Restoration Performance Evaluation}

To demonstrate the robustness and accuracy of our UIE method for color correction, we compare the color correction ability of 10 UIE methods on the Color-Checker7 dataset. The Color-Checker7 dataset contains 7 underwater images taken from a shallow swimming pool with different cameras.
Color checker is also photographed in each image. It provides a good path to demonstrate the robustness of our method to different imaging devices and the accuracy of color restoration. We follow Ancuti et al. \cite{Ancuti2018fusion} to employ CIEDE2000 \cite{Sharma2005color} to measure the relative differences between the corresponding color patches of ground-truth Macbeth Color Checker and the enhancement results of these comparison methods. The experimental results are shown in Tab. \ref{Tab:5} and Fig .\ref{figure:color_check}.

As in Tab. \ref{Tab:5}, for the cameras of Pentax W60, Pentax W80, Cannon D10, Fuji Z33, Panasonic TS1 and Olympus T6000, our U-shape Transformer obtains the lowest color dissimilarity. Moreover, our U-shape Transformer achieves the best average score. Such results demonstrate the superiority of our method for underwater color correction. It is worth mentioning that some comparable methods acquired lower score than that of the raw image, which reflected that those methods are incapable of recovering the real color and even break the inherent color.

As shown in Fig. \ref{figure:ablation}, the professional underwater camera (Fuji Z33) also inevitably introduces various color casts. Among all the UIE methods involved in the comparison, our U-shape Transformer achieves the highest CIEDE 2000 score, which means our UIE method has the best color correction ability. The results of UDCP and UIBLA are bluish, and Retinex has the problem of color distortion. UGAN and UIE-DAL suffer from low saturation and excessive reddish compensation. Although FUnIE and Ucolor 
could remove the color cast to a certain extent, there are still problems of low contrast and saturation.

\subsection{Ablation Study}
To prove the effectiveness of each component, we conduct a series of ablation studies on the Test-L400 and Test-U90. Four factors are considered including the CMSFFT, the SGFMT, the multi-scale gradient flow mechanism (MSG), and the multi-color space loss function (MCSL). 
\begin{table}[h]
	\centering
	\caption{Statistical results of ablation study on the Test-L400 and the Test-U90. The highest scores are marked in red.}
	\begin{tabular}{c||c|c|c|c|c}
		\hline
		\multicolumn{2}{c||}{\multirow{2}{*}{Models}} & \multicolumn{2}{c|}{Test-L400} & \multicolumn{2}{c}{Test-U90} \\ \cline{3-6} 
		\multicolumn{2}{c||}{}                        & PSNR           & SSIM          & PSNR           & SSIM         \\ \hline\hline
		\multicolumn{2}{c||}{BL}                      & 19.34          & 0.79          & 19.36          & 0.81         \\ 
		\multicolumn{2}{c||}{BL+CMSFFT}               & 22.47          & 0.88          & 21.72          & 0.86         \\ 
		\multicolumn{2}{c||}{BL+SGFMT}                & 21.78          & 0.86          & 21.36          & 0.87         \\ 
		\multicolumn{2}{c||}{BL+MSG}                  & 20.11          & 0.82          & 21.24          & 0.85         \\ 
		\multicolumn{2}{c||}{BL+MCSL}                 & 21.51          & 0.82          & 20.16          & 0.81         \\ 
		\multicolumn{2}{c||}{Full Model}              & \textcolor{red}{\textbf{24.16}}          & \textcolor{red}{\textbf{0.93}}          & \textcolor{red}{\textbf{22.91}}          & \textcolor{red}{\textbf{0.91}}         \\ \hline
	\end{tabular}
	\label{Tab:6}
\end{table}

\begin{figure}[h]
	\centering
	\includegraphics[width=1\linewidth]{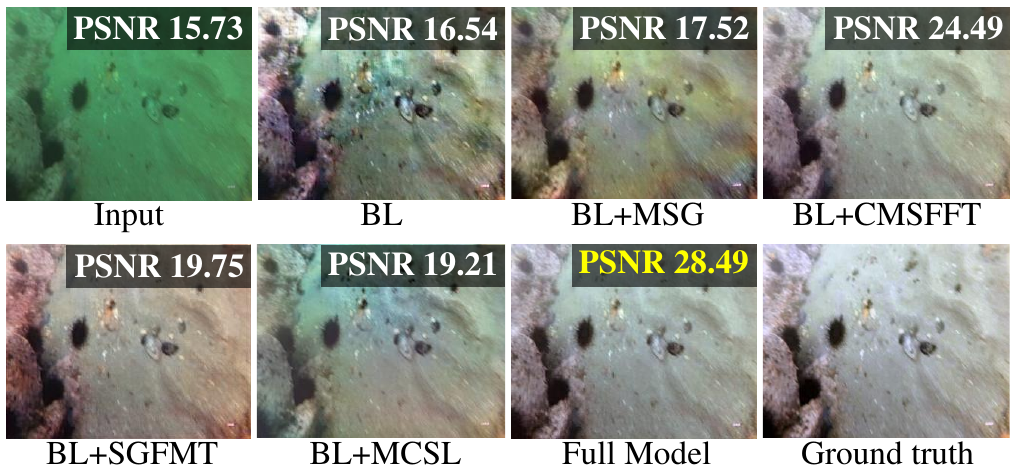}
	\caption{Visual comparison of the ablation study sampled from the Test-L400 dataset.}
	\label{figure:ablation}
\end{figure}

Experiments are all trained by Train-L. Statistical results are shown in Tab. \ref{Tab:6}, in which baseline model (BL) refers to \cite{isola2017image}, full models is the complete U-shape Transformer. In Tab. \ref{Tab:6}, our full model achieves the best quantitative performance on the two testing dataset, which reflects the effectiveness of the combination of CMSFFT, SGFMT, MSG, and MCSL modules. 
As in Fig .\ref{figure:ablation}, the enhancement result of the full model has the highest PSNR and best visual quality. The results of BL+MSG have less noise and artifacts than the BL module because the  MSG mechanism helps to reconstruct local details. Thanks to the multi-color space loss function, the overall color of BL+MCSL's result is close to the reference image.
The unevenly distributed visualization and artifacts in local areas of BL+MCSL are due to the lack of efficient attention guidance. Although the enhanced results of BL+CMSFFT and BL+SGFMT are evenly distributed, the overall color is not accurate. The investigated four modules have their particular functionality in the enhancement process, which integration could improve the overall performance of our network.

\section{Conclusions}
In this work, we released a large scale underwater image (LSUI) dataset, which contains 4279 real-world underwater images with more abundant underwater scenes (water types, lighting conditions and target categories) than existing underwater datasets \cite{Liu2020RealWorldUE,Li2019UIEB,Akkaynak2019seathru,li2017watergan}, and the corresponding clear images are generated as comparison references. We also provide the semantic  segmentation  map and medium  transmission  map for each raw underwater image. Besides, we reported an U-shape Transformer network for state-of-the-art UIE performance. The network's CMSFFT and SGFMT modules could solve the inconsistent attenuation issue of underwater images in different color channels and space regions, which has not been considered among existing methods. Extensive experiments validate the superior ability of the network to remove color artifacts and casts. Combined with the multi-color space loss function, the contrast and saturation of output images are further improved. Nevertheless, it is impossible to collect images of all the complicated scenes such as deep-ocean low-light scenarios. Therefore, we will introduce other general enhancement techniques such as low-light boosting \cite{Chen2018dark} for future work.

\section*{Acknowledgment}
This work was supported by the National Natural Science Foundation of China (61991451, 61971045, 62131003),
National Key Research and Development Program of China
(2020YFB0505601).

\ifCLASSOPTIONcaptionsoff
  \newpage
\fi



%
\bibliographystyle{IEEEtran}
\bibliography{bare_jrnl}

%








\end{document}